\documentclass[10pt,twocolumn,letterpaper]{article}

\usepackage{cvpr}
\usepackage{times}
\usepackage{epsfig}
\usepackage{graphicx}
\usepackage{amsmath}
\usepackage{amssymb}

\usepackage{enumitem}
\usepackage[nocompress]{cite}
\usepackage{verbatim}
\usepackage{multirow}
\usepackage{subfigure}
\usepackage{mathrsfs}
\usepackage{booktabs}  
\usepackage{threeparttable}
\usepackage{setspace}

\usepackage{colortbl}
\usepackage{color}

\usepackage{array}
\newcommand{\PreserveBackslash}[1]{\let\temp=\\#1\let\\=\temp}
\newcolumntype{C}[1]{>{\PreserveBackslash\centering}p{#1}}
\newcolumntype{R}[1]{>{\PreserveBackslash\raggedleft}p{#1}}
\newcolumntype{L}[1]{>{\PreserveBackslash\raggedright}p{#1}}

\usepackage[pagebackref=true,breaklinks=true,letterpaper=true,colorlinks,bookmarks=false]{hyperref}

\cvprfinalcopy 

\def\cvprPaperID{584} 

\ifcvprfinal\fi
\begin{document}
\definecolor{light_gray}{gray}{0.9}

\title{Camera Lens Super-Resolution}

\author{Chang Chen \hspace{0.4cm}
Zhiwei Xiong\thanks{Correspondence should be addressed to zwxiong@ustc.edu.cn}\hspace{0.45cm}
Xinmei Tian \hspace{0.4cm}
Zheng-Jun Zha \hspace{0.4cm}
Feng Wu\\
University of Science and Technology of China}

\maketitle

\begin{abstract}
Existing methods for single image super-resolution (SR) are typically evaluated with synthetic degradation models such as bicubic or Gaussian downsampling.
In this paper, we investigate SR from the perspective of camera lenses, named as CameraSR, which aims to alleviate the intrinsic tradeoff between resolution (R) and field-of-view (V) in realistic imaging systems.
Specifically, we view the R-V degradation as a latent model in the SR process and learn to reverse it with realistic low- and high-resolution image pairs.
To obtain the paired images, we propose two novel data acquisition strategies for two representative imaging systems (i.e., DSLR and smartphone cameras), respectively.
Based on the obtained \href{https://github.com/ngchc/CameraSR}{City100} dataset, we quantitatively analyze the performance of commonly-used synthetic degradation models, and demonstrate the superiority of CameraSR as a practical solution to boost the performance of existing SR methods.
Moreover, CameraSR can be readily generalized to different content and devices, which serves as an advanced digital zoom tool in realistic imaging systems.

\end{abstract}

\section{Introduction}
\label{sec:introduction}
Single image super-resolution (SR) is a typical inverse problem in computer vision. Generally, SR methods assume bicubic or Gaussian downsampling as the degradation model \cite{Multiple_Degradations}.
Based on this assumption, continuous progress has been achieved to restore a better high-resolution (HR) image from its low-resolution (LR) version, in terms of reconstruction accuracy \cite{ZSSR, VDSR, LapSRN, EDSR, DSC, DRRN, DBPN, RDN, RCAN, Xiong1, Xiong2} or perceptual quality \cite{SRGAN, SFT, EnhanceNet, SuperFAN, PIRM2018, Deng, PerceptualLoss}.
However, these synthetic degradation models may deviate from the ones in realistic imaging systems, which results in a significant deterioration on the SR performance \cite{BlindSR}.
To better simulate the challenging real-world conditions, additional factors including noise, motion blur, and compression artifacts are integrated to characterize the LR images in either a synthetic \cite{NTIRE2018} or a data-driven \cite{GAN_Degradation} manner.
These modified degradation models promote the SR performance of learning-based methods when the LR images indeed have corresponding degradations.

\begin{figure}[t]
\begin{center}
\begin{minipage}{0.435\linewidth}
  \centerline{\includegraphics[width=1\linewidth]{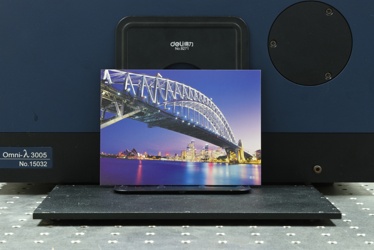}}
\end{minipage}
\hfill
\begin{minipage}{0.551\linewidth}
  \centerline{\includegraphics[width=1\linewidth]{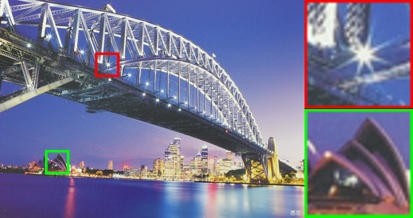}}
\end{minipage}
\vfill
\vspace{0.03cm}
\begin{minipage}{0.435\linewidth}
  \centerline{\includegraphics[width=1\linewidth]{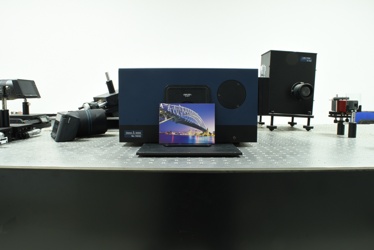}}
  \vspace{-0.05cm}
  \fontsize{9}{10.8}\selectfont
  \centerline{(a)}
\end{minipage}
\hfill
\begin{minipage}{0.551\linewidth}
  \centerline{\includegraphics[width=1\linewidth]{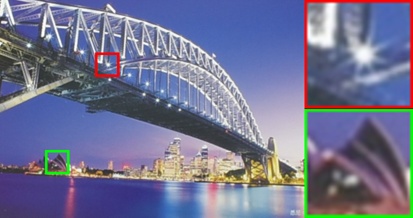}}
  \vspace{-0.05cm}
  \fontsize{9}{10.8}\selectfont
  \centerline{(b)}
\end{minipage}
\end{center}
\vspace{-0.20cm}
\caption{(a) Resolution-FoV (R-V) degradation. Zooming out the optical lens in a DSLR camera, the FoV is enlarged at the cost of resolution loss. (b) Aligned realistic LR-HR image pair after rectification. LR image is displayed after interpolation for a side-by-side comparison. (Bicubic interpolation is used throughout this paper unless noted otherwise.)}
\vspace{-0.20cm}
\label{fig:raw2calib}
\end{figure}

In this paper, we investigate SR from the perspective of camera lenses, named as CameraSR, which aims to alleviate the intrinsic tradeoff between resolution (R) and field-of-view (FoV, V for short hereafter) in realistic imaging systems. An instance of the R-V tradoff is shown in Fig.~\ref{fig:raw2calib}(a). When zooming out the optical lens in a DSLR camera, the obtained image has a larger FoV but loses details on subjects; when zooming in the lens, the details of subjects show up at the cost of a reduced FoV.
This R-V tradeoff also applies to cameras with fixed focal lenses such as those on smartphones when the shooting distance changes. Inspired by learning-based single image SR, we view the \textit{R-V degradation} (i.e., resolution loss due to enlarged FoV) as a latent model in the SR process and learn to reverse it with a number of LR-HR image pairs. Specifically, we define a subject captured at a long focal length or a short distance as the HR ground truth, and the same one captured at a short focal length or a long distance as its paired LR observation.

To obtain such paired images, we first use a DSLR camera mounted on a tripod with a zoom lens. To avoid the out-of-focus blur, we adopt a small aperture size and capture 100 city scenes printed on postcards as the subjects which can be well focused at different focal lengths.
In practice, however, several issues due to the mechanical zoom prohibit the direct use of the captured raw data, including spatial misalignment, intensity variation, and color mismatching. After addressing these issues through an elaborate data rectification pipeline, we build a dataset consisting of 100 aligned image pairs named ``City100''. An example is shown in Fig.~\ref{fig:raw2calib}(b).
Following the same pipeline, we then obtain a variant of City100 by using a smartphone camera mounted on a translation stage with a fixed focal lens. The City100 dataset, together with its smartphone version, characterizes the R-V degradation in two representative realistic imaging systems.

\begin{figure}[t]
\begin{center}
\begin{minipage}{0.325\linewidth}
  \centerline{\includegraphics[width=1\linewidth]{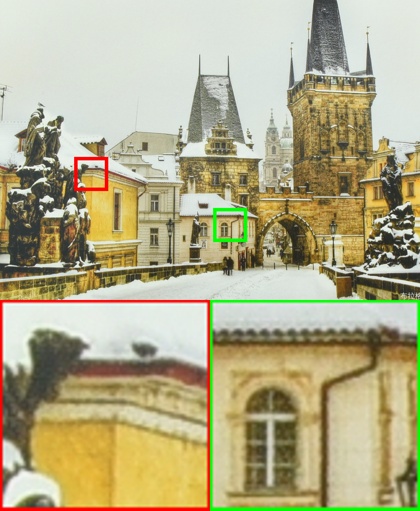}}
  \fontsize{7}{8.4}\selectfont
  \centerline{(a) HR Ground Truth}
  \centerline{PSNR}
\end{minipage}
\hfill
\begin{minipage}{0.325\linewidth}
  \centerline{\includegraphics[width=1\linewidth]{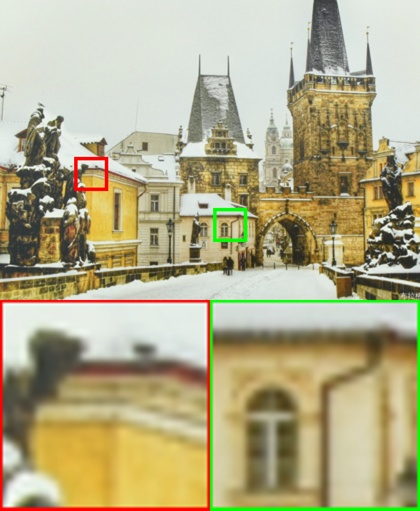}}
  \fontsize{7}{8.4}\selectfont
  \centerline{(b) Bicubic Downsampling}
  \centerline{28.56 dB}
\end{minipage}
\hfill
\begin{minipage}{0.325\linewidth}
  \centerline{\includegraphics[width=1\linewidth]{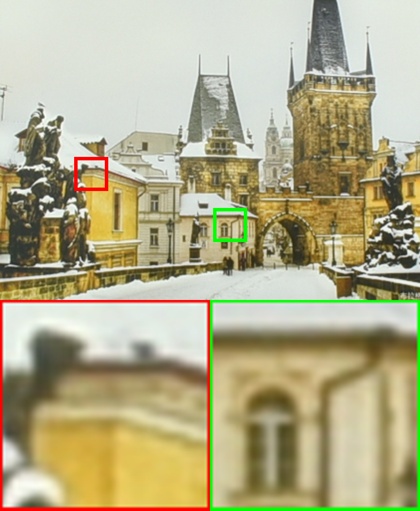}}
  \fontsize{7}{8.4}\selectfont
  \centerline{(c) R-V Degradation}
  \centerline{27.22 dB}
\end{minipage}
\end{center}
\caption{Visual comparison between the LR image with bicubic downsampling and the realistic LR image with R-V degradation (both are displayed after interpolation). The latter loses more information than the former in visual compared with the HR ground truth, which is also quantitatively verified by PSNR.}
\label{fig:comparison_bic_real}
\vspace{-0.15cm}
\end{figure}

\begin{figure}
\fontsize{9}{10.8}\selectfont
\begin{minipage}{0.24\linewidth}
  \centerline{\includegraphics[width=1\linewidth]{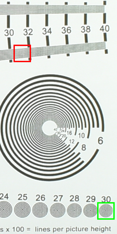}}
\end{minipage}
\hfill
\begin{minipage}{0.745\linewidth}
\begin{minipage}{0.32\linewidth}
  \centerline{\includegraphics[width=1\linewidth]{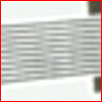}}
\end{minipage}
\hfill
\hspace{-0.15cm}
\begin{minipage}{0.32\linewidth}
  \centerline{\includegraphics[width=1\linewidth]{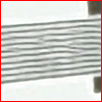}}
\end{minipage}
\hfill
\hspace{-0.15cm}
\begin{minipage}{0.32\linewidth}
  \centerline{\includegraphics[width=1\linewidth]{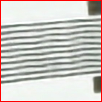}}
\end{minipage}
\vfill
\begin{minipage}{0.32\linewidth}
  \centerline{\includegraphics[width=1\linewidth]{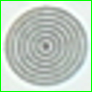}}
\end{minipage}
\hfill
\hspace{-0.15cm}
\begin{minipage}{0.32\linewidth}
  \centerline{\includegraphics[width=1\linewidth]{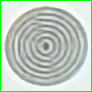}}
\end{minipage}
\hfill
\hspace{-0.15cm}
\begin{minipage}{0.32\linewidth}
  \centerline{\includegraphics[width=1\linewidth]{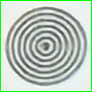}}
\end{minipage}
\end{minipage}
\vfill
\vspace{0.1cm}
\begin{minipage}{0.24\linewidth}
  \centerline{(a)}
\end{minipage}
\hfill
\begin{minipage}{0.745\linewidth}
\begin{minipage}{0.32\linewidth}
  \centerline{(b)}
\end{minipage}
\hfill
\begin{minipage}{0.32\linewidth}
  \centerline{(c)}
\end{minipage}
\hfill
\begin{minipage}{0.32\linewidth}
  \centerline{(d)}
\end{minipage}
\end{minipage}
\vspace{-0.18cm}
\caption{An example to show the performance deterioration due to improper degradation modeling (bicubic downsampling here). (a) An image captured by a DSLR camera. (b) Interpolated result. (c) SR result using VDSR \cite{VDSR} trained under bicubic downsampling. (d) SR result using VDSR trained under R-V degradation.}
\vspace{-0.06cm}
\label{fig:comparison_toy}
\end{figure}

Based on City100, we conduct a quantitative analysis on the commonly-used synthetic degradation models, in terms of both LR observations and SR results. Take the bicubic downsampling as an example, due to the underestimation of R-V degradation (as shown in Fig.~\ref{fig:comparison_bic_real}), it results in a significant deterioration on the SR performance (as shown in Fig.~\ref{fig:comparison_toy}). This analysis validates the importance of degradation modeling for the resolution enhancement in realistic imaging systems.
Observing the disadvantage of synthetic degradation models, we propose CameraSR as a practical solution to boost the performance of existing SR methods, by learning the R-V degradation from City100. Comprehensive experiments demonstrate that CameraSR achieves a significant improvement of SR results compared with those using synthetic degradation models.

More importantly, we demonstrate that CamereSR has a favorable capability of generalization in terms of both content and device. Specifically, an SR network trained on City100 can be readily generalized to other scene content, as well as to other type of devices belonging to the same category of imaging systems (e.g., from Nikon to other DSLRs and from iPhone to other smartphones). By effectively alleviating the R-V tradeoff or even breaking the physical zoom ratio of an optical lens in realistic imaging systems, CameraSR could find a wide application in practice as an advanced digital zoom tool.

Contributions of this paper are summarized as follows:
\vspace{-0.18cm}
\begin{itemize} \itemsep -1.1pt

\item A new perspective, i.e., R-V degradation of camera lenses, for SR modeling in realistic imaging systems.

\item Two novel strategies for acquiring LR-HR image pairs as in \href{https://github.com/ngchc/CameraSR}{City100} to characterize the R-V degradation under DSLR and smartphone cameras, respectively.

\item Quantitative analysis on the commonly-used synthetic degradation models using realistic data.

\item An effective solution, i.e., CameraSR, to promote existing learning-based SR methods in realistic imaging systems.

\end{itemize}

\section{Related Work}
Recent years have seen a remarkable improvement in single image SR. To promote the reconstruction accuracy, increasingly more learning-based methods adopt the convolutional neural network (CNN) following the seminal work of SRCNN \cite{SRCNN}.
For instance, Kim \etal proposed VDSR \cite{VDSR} which deepens the network for accuracy with the residual learning. Lai \etal proposed LapSRN \cite{LapSRN} which improves the SR results at large scale factors with the Laplacian pyramid structure. Furthermore, various mechanisms have been integrated into the network design to advance the SR performance, such as sparsity \cite{SCN}, contiguous memory \cite{RDN}, deep supervision \cite{DSC}, recursion \cite{DRCN, DRRN}, back-projection \cite{DBPN}, information distillation \cite{IDN}, and attention \cite{RCAN}.
Different from the above methods, Ledig \etal proposed SRGAN \cite{SRGAN} which is optimized for perceptual quality instead of reconstruction accuracy. Along this line, Sajjadi \etal proposed EnhanceNet \cite{EnhanceNet} which promotes the quality of texture synthesis with a perceptual loss. Wang \etal proposed SFTGAN \cite{SFT} which integrates a spatial feature transform layer into GAN \cite{GAN} to further enhance the SR performance.
However, most existing learning-based methods adopt a synthetic degradation model (e.g., bicubic or Gaussian downsampling) when formulating the SR problem, which hinders their performance in realistic imaging systems with much more complicated degradation.

There are a few works that involve realistic degradation modeling for single image SR. For instance, Timofte \etal introduced more degradation operators into the bicubic-downsampled LR images, including motion blur and Poisson noise \cite{NTIRE2018}. Bulat \etal defined the LR face images with the low-quality assumptions (e.g., noise, blur, and compression artifacts) and trained a GAN \cite{GAN} to learn the degradation process \cite{GAN_Degradation}. On the other hand, as a self-similarity based method, Michaeli and Irani adaptively estimated the degradation model relying on the inherent recurrence of the input image \cite{BlindSR}. Shocher \etal further optimized an image-specific CNN with examples solely extracted from the input image \cite{ZSSR}.

Different from the above approaches, our proposed CameraSR models the R-V degradation from the perspective of camera lenses. The estimation of R-V degradation neither relies on the low-quality assumptions nor the inherent recurrence of LR images.
Instead, it is characterized by the samples captured with realistic imaging systems. Such a degradation modeling is inspired by the prior work for realistic image denoising \cite{DND}, where a subject captured at a high ISO value is defined noisy and the same one captured at a low ISO value is defined clean.
We extend this definition to the SR scenario, which addresses the key challenge of obtaining realistic LR-HR image pairs. Note that the focus of this paper is not the network design. For the comparison purpose, we adopt VDSR \cite{VDSR} and SRGAN \cite{SRGAN} as two representative embodiments to demonstrate the effectiveness and generalizability of CameraSR, which can be replaced with any CNN-based methods.

\section{Problem Formulation}

Consider again taking photos using a DSLR camera with an optical zoom lens. Zooming out the lens derives a larger FoV at the cost of resolution loss on the subject. Denote this R-V degradation as $D_{RV}(\cdot)$, our goal is to obtain a function $S(\cdot)$ that reverses $D_{RV}(\cdot)$ for realistic image SR. This problem can be formulated as
\begin{equation}
\hat{X}=S(D_{RV}(X)),
\end{equation}
where $X$ denotes the original image and $\hat{X}$ denotes the super-resolved one. Compared with previous SR formulations, the only difference lies in the modeling for the degradation process. For instance, the bicubic downsampling $D_{Bic}(\cdot)$ formulates the SR problem as $\hat{X}=S(D_{Bic}(X))$ and the Gaussian downsampling as $\hat{X}=S(D_{Gau}(X))$. For the more complicated degradation model imposed in \cite{NTIRE2018}, it is $\hat{X}=S(D_{Blur}(D_{Bic}(X))+v)$, where $D_{Blur}(\cdot)$ denotes a blurring operator and $v$ denotes a certain kind of noise.

Unlike the synthetic degradation models as mentioned above, it is difficult to derive an analytic expression for $D_{RV}(\cdot)$. Inspired by learning-based SR, we view the R-V degradation as a latent model $\hat{D}_{RV}(\cdot)$ in the SR process and directly learn the parametric SR function $S_\Theta(\cdot)$ with $N$ pairs of realistic LR ($\mathbf{Y}=\{Y_1, Y_2, ..., Y_N\}$) and HR ($\mathbf{X}=\{X_1, X_2, ..., X_N\}$) samples, which can be represented as
\begin{equation}
\hat{X}=S_\Theta(\hat{D}_{RV}(X)),
\end{equation}
where $\hat{D}_{RV}(\cdot)$ is subject to $\mathbf{Y}=\hat{D}_{RV}(\mathbf{X})$. With the increase of the number of samples $N$, we have $\hat{D}_{RV}(\cdot)\rightarrow D_{RV}(\cdot)$. Then, $S_\Theta(\cdot)$ can be optimized with a loss function $\mathcal{L}(\cdot)$ as
\begin{equation}
\label{eq:s}
\min_{\Theta}\dfrac{1}{n}\sum_{i=1}^{n}\mathcal{L}(X_i-S_\Theta(Y_i)),
\end{equation}
where $\Theta$ denotes a set of trainable parameters and $n$ denotes the size of mini-batch when optimizing $\Theta$ with the stochastic gradient descent algorithm. 

This is the main idea of CameraSR, which will be detailed in Sec.~\ref{sec:comp_sr}. While the problem formulation is quite intuitive, the key challenge is, how to obtain the LR-HR image pairs in realistic imaging systems?

\section{Data Acquisition}
\subsection{DSLR imaging system}
\label{sec:postprocessing}

To capture the realistic LR-HR image pairs, we use a Nikon D5500 camera mounted on a tripod with a Zoom-Nikkor lens, whose focal length ranges from 18mm to 55mm. We define an image captured at 55mm focal length as the HR ground truth and the one captured at 18mm focal length as the LR observation. To alleviate the influence of noise, the ISO value is set to the lowest level. The other settings such as white balance and aperture size are fixed for each capture.
In practice, however, we observe several issues for prohibiting the direct use of the captured raw data, including spatial misalignment, intensity variation, and color mismatching.
It is probably due to the fact that the change of focal length is a mechanical process which cannot be ideally controlled. It thus results in slight dithering of the camera body as well as the exposure configuration. To address these issues, we elaborate a data rectification pipeline.

\begin{figure}
\begin{center}
  \includegraphics[width=1\linewidth]{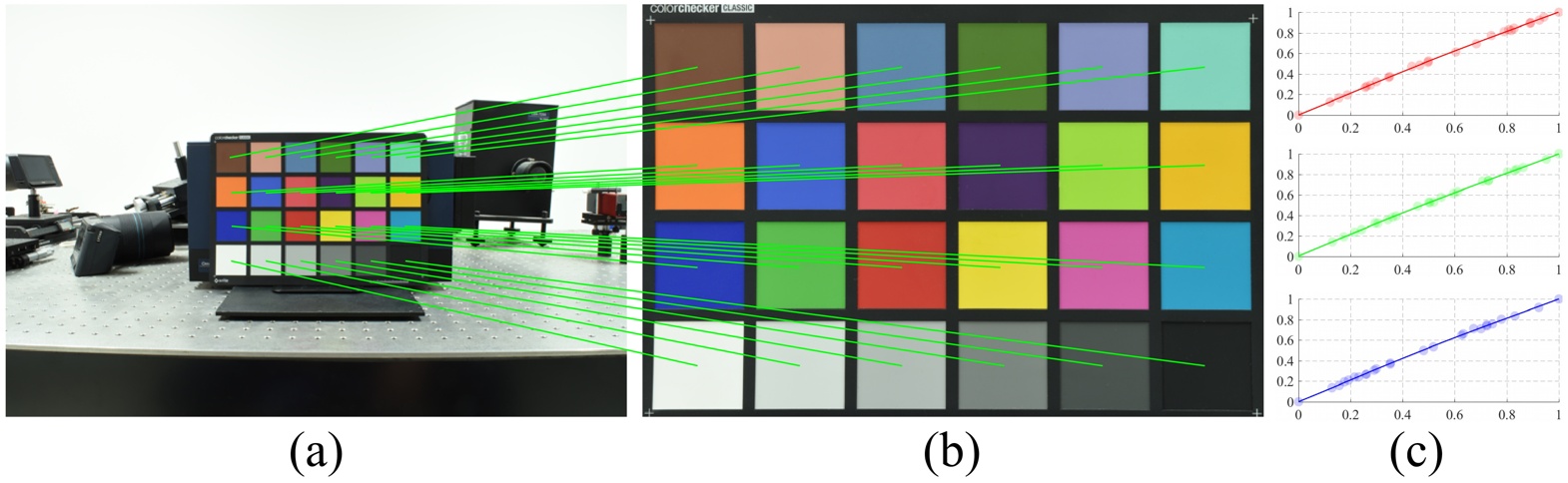}
\end{center}
\vspace{-0.2cm}
\caption{Color calibration. The mean values obtained from each color blocks are adopted to fit three polynomial curves (c) for color calibration, from the LR observation (a) to its HR ground truth (b).}
\vspace{-0.1cm}
\label{fig:calibration}
\end{figure}

First, we model the spatial misalignment as a global 2D translation inspired by \cite{DPED}. Specifically, we compute and match SIFT key-points \cite{SIFT} between the HR images and the interpolated LR ones.
Then, the matched coordinates are used to estimate a homography using RANSAC \cite{RANSAC}. Having the translation parameters, we shift the LR images through interpolation to obtain the aligned results. Note that the interpolation will introduce some smoothing effects, but not critical for the already interpolated LR images which contain relatively fewer high frequencies. We avoid shifting the HR images since they contain a lot of desired details.
Second, we model the intensity variation as a bias in the DC component of an image and estimate it by averaging the pixel intensities in the whole image. Then, we use the estimated bias to compensate this variation.
Third, we model the color mismatching as a parametric non-linear mapping and fit it with polynomial parameters for calibration by leveraging a color checkerboard, as shown in Fig.~\ref{fig:calibration}.
Specifically, we collect and average pixel values in each block from the color checkerboard to obtain paired samples from the LR observation to its HR ground truth. Then, we fit three polynomial curves for R, G, and B channels using the collected samples, respectively. Finally, we map pixels in LR observations using the obtained polynomial curves.

After the above data rectification, we build a City100 dataset using the DSLR camera, in which 100 city scenes printed on high-quality postcards are adopted as the subjects. The plane shape of postcards guarantees that the whole image can be well focused under a small aperture size at both long and short focal lengths, which avoids the out-of-focus blur. 
The resolution of final HR images in City100 is $1218\times870$, which is 2.9 times of the LR ones. Images from City100 have diverse colors and contents, which facilitate leaning-based SR. An overview of the City100 dataset is shown in the supplementary document.

\subsection{Smartphone imaging system}

\begin{figure}
\begin{center}
  \includegraphics[width=1.0\linewidth]{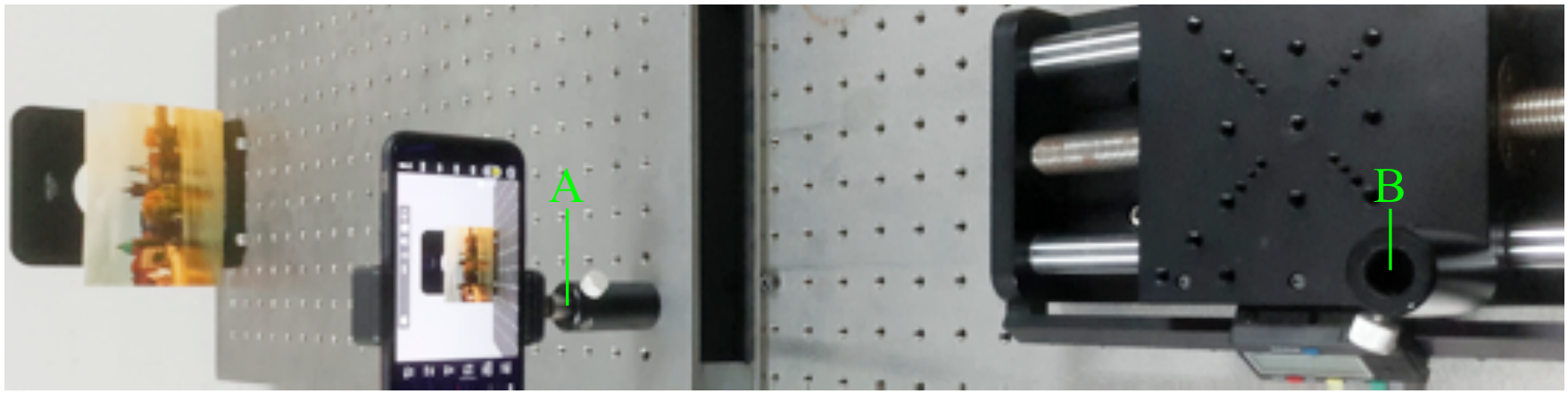}
\end{center}
\vspace{-0.1cm}
\caption{Acquisition strategy for the smartphone version of City 100. Translating the smartphone away from a subject (from A to B), the effective resolution is decreased due to the enlarged FoV (R-V degradation).}
\vspace{-0.07cm}
\label{fig:spac}
\end{figure}

Different from the zoom lenses in professional DSLR cameras, commodity smartphone cameras are generally equipped with prime lenses whose focal length cannot change. In this sense, the realistic degradation modeling is even more meaningful to smartphones, where CameraSR can serve as a powerful digital zoom tool. However, limited by the fixed focal lens, LR-HR image pairs for smartphone cameras cannot be captured with the same strategy as for DSLR cameras. Alternatively, we develop another strategy for obtaining the smartphone version of City100, as shown in Fig.~\ref{fig:spac}.
An iPhone X mounted on a translation stage is used for data acquisition, and the position of iPhone relative to the translation stage can be precisely adjusted. We define an image captured at a short distance as the HR ground truth, and the one captured at a long distance as the LR observation. To avoid the ``intelligent'' exposure configuration by the smartphone itself, we use the ProCam\footnote{\url{https://www.procamapp.com}} software to manually control the settings such as ISO, white balance, exposure time and so on. The data rectification pipeline for smartphone is similar to that for DSLR as detailed in Sec.~\ref{sec:postprocessing}. In addition, considering that smartphone images have notably heavier noise than DSLR images due to the much smaller sensor size, we repeat the capture of each scene 20 times and average the resulting images to alleviate the influence of noise. The resolution of final HR images is 2.4 times of the LR ones.

It is worth mentioning that, the City100 dataset and its smartphone version are obtained by two representative realistic imaging systems, i.e., DSLR and smartphone. Although two specific devices, i.e., Nikon D5500 and iPhone X are used here, the trained CameraSR network has a favorable capability of generalization and can be readily applied to different devices belonging to the same category of imaging systems (as detailed in Sec.~\ref{sec:generalization}).

\section{Analysis on Degradation Models}

In this section, our goal is to quantitatively analyze the performance of commonly used synthetic degradation models $D_{Bic}(\cdot)$ and $D_{Gau}(\cdot)$, in comparison with the realistic R-V degradation $D_{RV}(\cdot)$ based on the paired samples from our developed City100 dataset. Since $D_{RV}(\cdot)$ has not an analytic expression, it is difficult to conduct direct comparisons between them. Thus, we turn to the corresponding LR observations and SR results for quantitative comparisons.

\begin{figure}
\begin{center}
\begin{minipage}{0.55\linewidth}
  \includegraphics[width=1\linewidth]{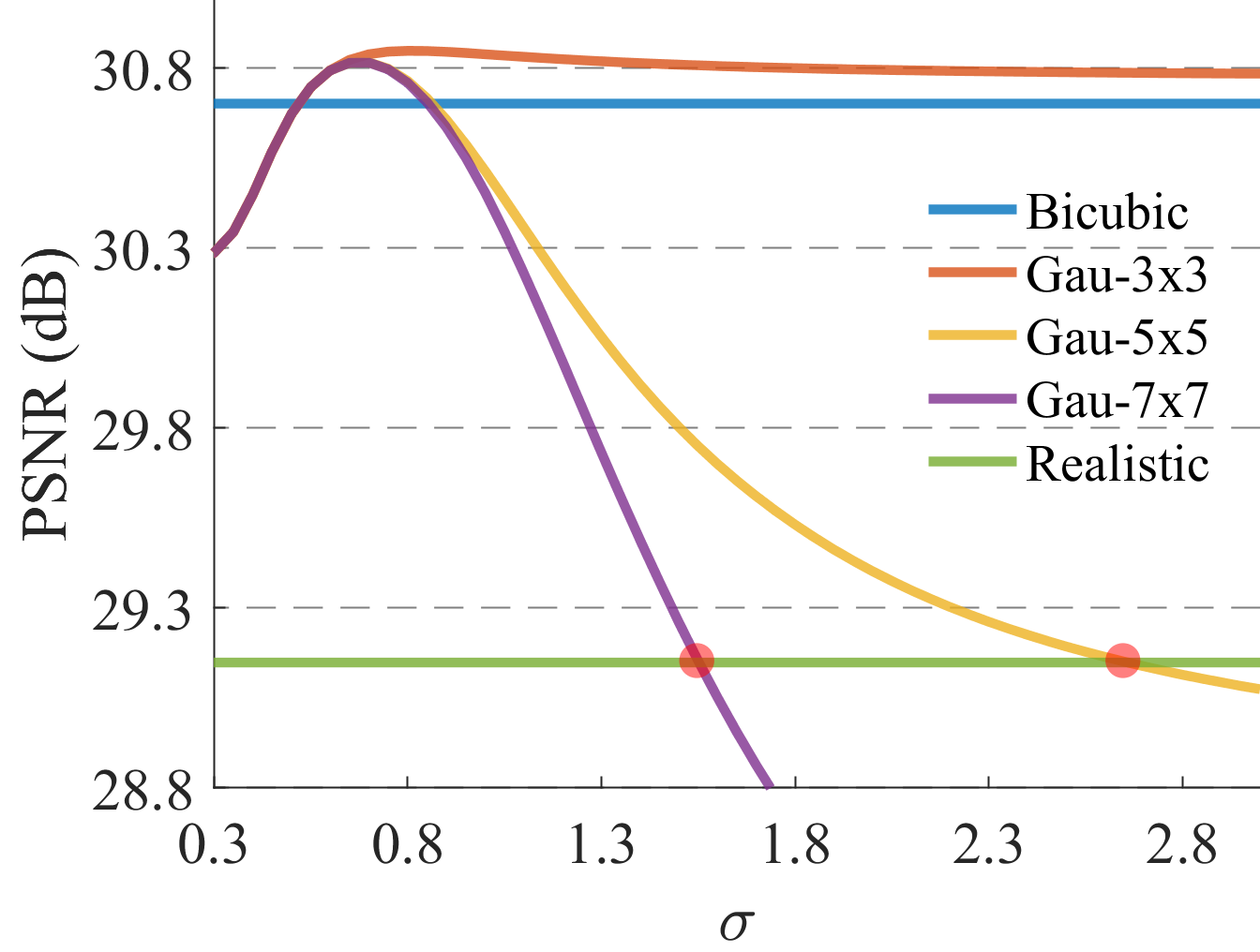}
  \vspace{-0.2cm}
  \centerline{(a)}
\end{minipage}
\begin{minipage}{0.41\linewidth}
  \includegraphics[width=1\linewidth]{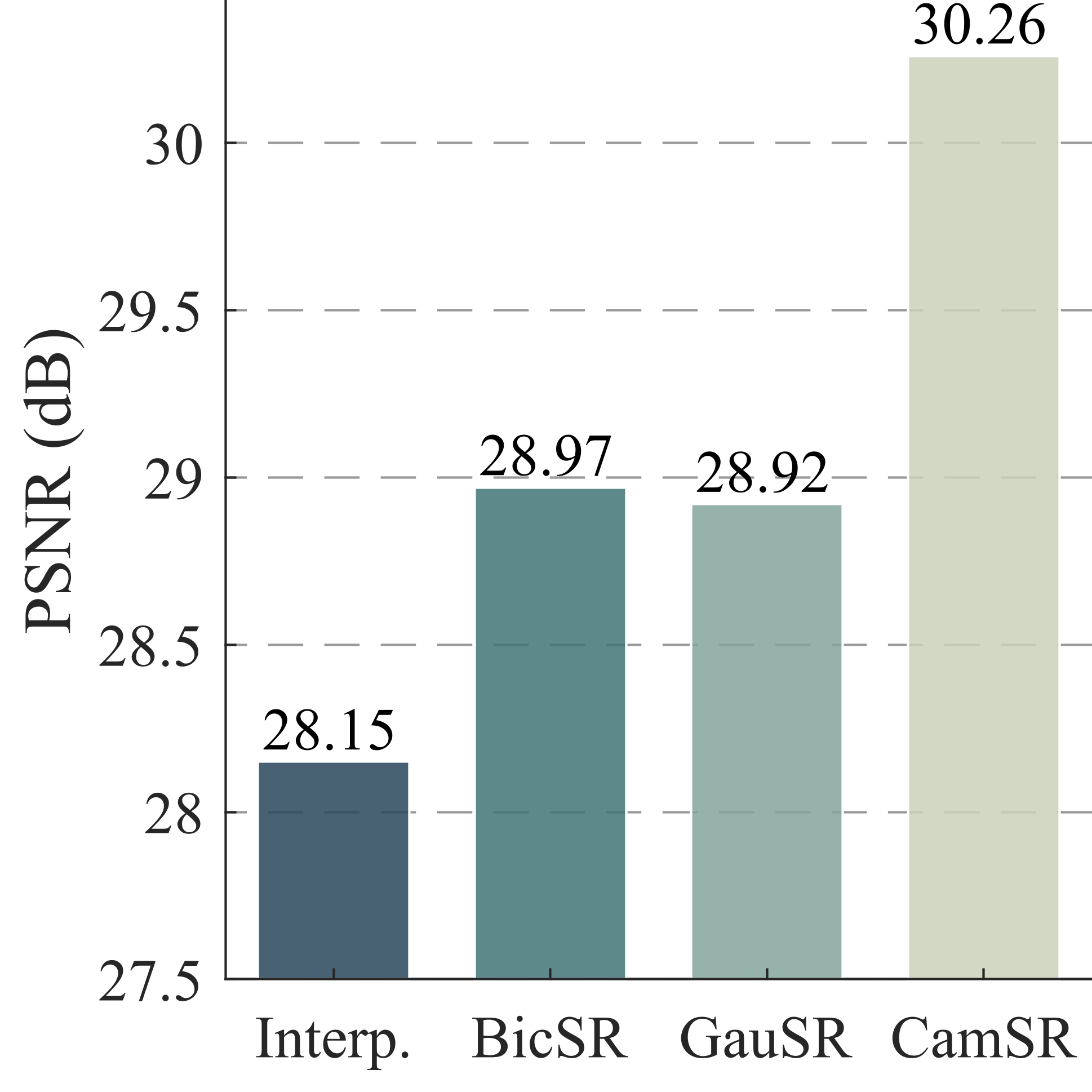}
  \vspace{-0.2cm}
  \centerline{(b)}
\end{minipage}
\end{center}
\caption{Analysis on the synthetic degradation models. (a) Investigation on the LR observations from City100. The PSNR is calculated between an interpolated LR image and its HR ground truth. (b) Investigation on the SR results from test set (as shown in Fig.~\ref{fig:city100_testset}). VDSR \cite{VDSR} is adopted as a representative network for BicubicSR, GaussianSR, and CameraSR. Although the Gaussian downsampling matches the degradation level of the realistic LR observation at the red points, the reconstruction accuracy of GaussianSR still has a gap compared with CameraSR. It reveals the disadvantage of synthetic degradation models.}
\vspace{-0.4cm}
\label{fig:comparison_lr_sr}
\end{figure}

\begin{table*}[t]
\fontsize{8}{9.6}\selectfont
\begin{center}
\begin{threeparttable}
\begin{tabular}{L{1.4cm}C{3.45cm}C{3.45cm}C{3.45cm}C{3.45cm}}
\toprule[1.2pt]
\multirow{2}{*}{Test image} & Interpolated LR & BicubicSR & GaussianSR & CameraSR \\
					\cmidrule(lr){2-2} \cmidrule(lr){3-3} \cmidrule(lr){4-4} \cmidrule(lr){5-5}
& PSNR / SSIM / Ma's / VGG & PSNR / SSIM / Ma's / VGG & PSNR / SSIM / Ma's / VGG & PSNR / SSIM / Ma's / VGG \\
\cmidrule(r){1-1} \cmidrule(){2-5}
St. Petersb. & 28.74 / 0.8630 / 3.58 / 0.8543 & 29.69 / 0.8874 / 5.05 / 0.7756 & 29.61 / 0.8934 / 6.16 / 0.7019 & 31.00 / 0.9116 / 6.58 / 0.4791 \\
Dubai         & 30.21 / 0.8443 / 3.37 / 0.5650 & 30.91 / 0.8599 / 4.73 / 0.4193 & 30.71 / 0.8603 / 5.86 / 0.3856 & 31.94 / 0.8788 / 6.74 / 0.3390 \\
Venice        & 26.52 / 0.7317 / 3.58 / 0.9654 & 27.25 / 0.7686 / 4.43 / 0.8254 & 27.21 / 0.7813 / 5.93 / 0.7798 & 28.19 / 0.8062 / 6.71 / 0.6167 \\
Rome         & 30.65 / 0.8654 / 3.60 / 0.3825 & 31.45 / 0.8806 / 4.77 / 0.3625 & 30.99 / 0.8768 / 6.17 / 0.3525 & 33.04 / 0.9039 / 6.68 / 0.2891 \\
New York  & 24.62 / 0.7520 / 3.83 / 1.1808 & 25.55 / 0.7921 / 4.85 / 1.1528 & 26.06 / 0.8113 / 5.85 / 1.1345 & 27.14 / 0.8416 / 6.76 / 0.8381 \\
\cmidrule(r){1-1} \cmidrule(){2-5}
Average     & 28.15 / 0.8113 / 3.59 / 0.7896 & 28.97 / 0.8377 / 4.77 / 0.7071 & 28.92 / 0.8446 / 5.99 / 0.6709 & 30.26 / 0.8684 / 6.69 / 0.5124 \\
\bottomrule[1.2pt]
\end{tabular}
\end{threeparttable}
\end{center}
\vspace{-0.1cm}
\caption{Quantitative results of SR on the five test images from City100 (as shown in Fig.~\ref{fig:city100_testset}). PSNR and SSIM \cite{SSIM} (the higher, the better) are adopted for the evaluation of reconstruction accuracy (VDSR \cite{VDSR} network). Ma's metric \cite{Ma-Metric} (the higher, the better) and the VGG metric (the lower, the better) are adopted for the evaluation of perceptual quality (SRGAN \cite{SRGAN} network). We denote the Euclidean distance between SR results and ground truth in the feature space of a trained VGG-19 \cite{VGG} network as the VGG metric $(\times10e^4)$ \cite{Zhang_2018_CVPR}.} 
\label{tab}
\vspace{-0.3cm}
\end{table*}

\subsection{LR observation}

Given an HR image $X$ from City100, the LR observations are obtained by $D_{Bic}(X)$, $D_{Gau}(X)$, and $D_{RV}(X)$ (i.e., the paired $Y$ from City100), respectively. As demonstrated in Fig.~\ref{fig:comparison_bic_real}, $D_{Bic}(\cdot)$ underestimates the degradation level of $D_{RV}(\cdot)$, which results in a significant deterioration on the SR performance as shown in Fig.~\ref{fig:comparison_toy}.
Besides $D_{Bic}(\cdot)$, we further investigate $D_{Gau}(\cdot)$. In practice, the Gaussian downsampling first blurs $X$ with a Gaussian filter and then conducts pixel decimation at designated scale factors. To match the scale factor of samples from City100, we adapt $D_{Gau}(\cdot)$ for $\times2.9$ downsampling by first interpolating an image $X$ 3/2.9 times followed by a $\times 3$ decimation.
In contrast to the bicubic downsampling, the Gaussian downsampling is more flexible as its kernel size $k \times k$ and standard deviation $\sigma$ can be manually controlled. Here, we consider an ideal condition when the degradation level of $D_{Gau}(X)$ matches $D_{RV}(X)$ in terms of the LR observation.
To this end, we traverse $k$ and $\sigma$ as shown in Fig.~\ref{fig:comparison_lr_sr}(a). After interpolating $D_{Gau}(X)$ and $D_{RV}(X)$ to the same resolution
as $X$, we calculate the mean PSNR between them on City100 and find two matched parameters at the red points (with $k_1=5$, $\sigma_1=2.65$ and $k_2=7$, $\sigma_2=1.55$), which are adopted as the representatives of $D_{Gau}(\cdot)$.

\subsection{SR result}
\label{sec:comp_sr}

\begin{figure}
\begin{center}
\begin{minipage}{0.192\linewidth}
  \includegraphics[width=1\linewidth]{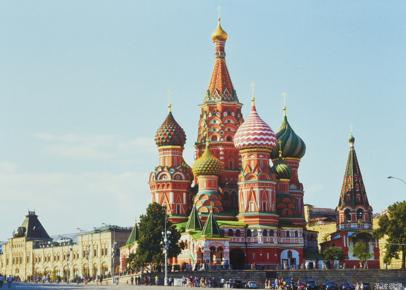}
  \fontsize{7}{8.4}\selectfont
  \vspace{-0.05cm}
  \centerline{St. Petersburg}
\end{minipage}
\hfill
\begin{minipage}{0.192\linewidth}
  \includegraphics[width=1\linewidth]{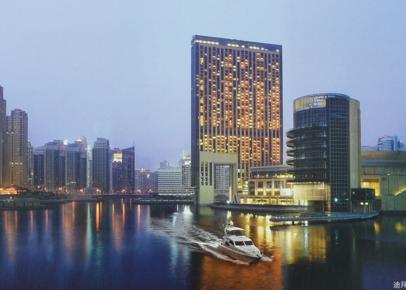}
  \fontsize{7}{8.4}\selectfont
  \centerline{Dubai}
\end{minipage}
\hfill
\begin{minipage}{0.192\linewidth}
  \includegraphics[width=1\linewidth]{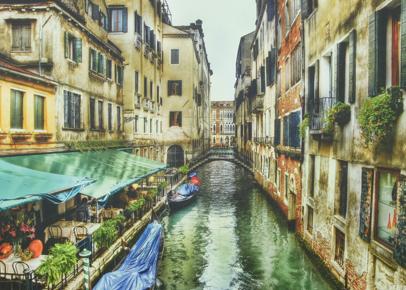}
  \fontsize{7}{8.4}\selectfont
  \centerline{Venice}
\end{minipage}
\hfill
\begin{minipage}{0.192\linewidth}
  \includegraphics[width=1\linewidth]{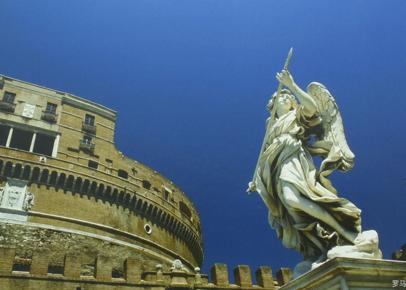}
  \fontsize{7}{8.4}\selectfont
  \centerline{Rome}
\end{minipage}
\hfill
\begin{minipage}{0.192\linewidth}
  \includegraphics[width=1\linewidth]{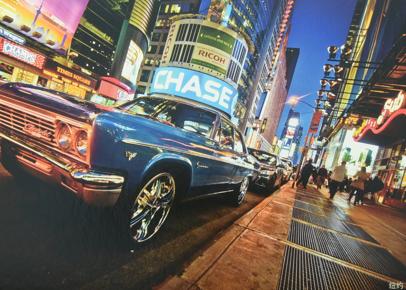}
  \fontsize{7}{8.4}\selectfont
  \centerline{New York}
\end{minipage}
\hfill
\end{center}
\caption{Thumbnails of the five test images from City100.}
\label{fig:city100_testset}
\end{figure}

Obtained the LR observations, we then evaluate the performance of different degradation models on the SR results, by comparing $S(D_{Bic}(X))$, $S(D_{Gau}(X))$, and $S(D_{RV}(X))$ to the ground truth $X$. We name the corresponding SR processes as BicubicSR, GaussianSR, and CameraSR for short, respectively.
To train an SR network, we split City100 into two parts: 5 selected pairs for test (as shown in Fig.~\ref{fig:city100_testset}) and the other 95 pairs for training. Among the training set, 5 images are used for validation. For the baseline network, we adopt two representative CNN architectures considering the perception-distortion
tradeoff reported in \cite{PD-Tradeoff}. For reconstruction accuracy, we adopt the VDSR network \cite{VDSR} with a mean square loss
\begin{equation}
\mathcal{L}_{MSE} = ||S_\Theta(D(x))-x||_2^2,
\end{equation}
where $x$ denotes an image patch cropped from $X$ on City100, $D(\cdot)$ denotes the a certain degradation model, and $S_\Theta(\cdot)$ denotes the parametric SR network.

\begin{figure*}
\fontsize{8}{9.6}\selectfont
\begin{center}
\begin{minipage}{1.0\linewidth}
\begin{minipage}{0.195\linewidth}
  \centerline{\includegraphics[width=1\linewidth]{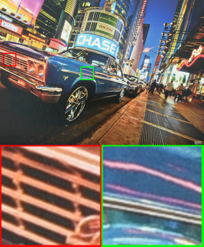}}
  \vspace{-0.03cm}
  \centerline{Ground Truth HR}
  \centerline{(PSNR / SSIM)}
\end{minipage}
\hfill
\begin{minipage}{0.195\linewidth}
  \centerline{\includegraphics[width=1\linewidth]{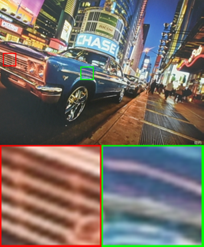}}
  \vspace{-0.03cm}
  \centerline{Interpolated LR}
  \centerline{(24.62 / 0.7520)}
\end{minipage}
\hfill
\begin{minipage}{0.195\linewidth}
  \centerline{\includegraphics[width=1\linewidth]{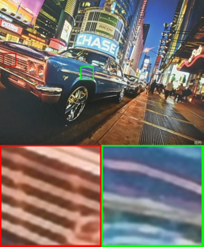}}
  \vspace{-0.03cm}
  \centerline{BicubicSR}
  \centerline{(25.55 / 0.7921)}
\end{minipage}
\hfill
\begin{minipage}{0.195\linewidth}
  \centerline{\includegraphics[width=1\linewidth]{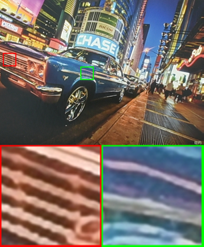}}
  \vspace{-0.03cm}
  \centerline{GaussianSR}
  \centerline{(26.06 / 0.8113)}
\end{minipage}
\hfill
\begin{minipage}{0.195\linewidth}
  \centerline{\includegraphics[width=1\linewidth]{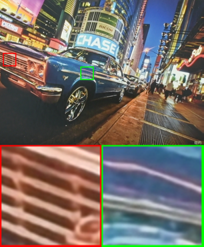}}
  \vspace{-0.03cm}
  \centerline{CameraSR}
  \centerline{(27.14 / 0.8416)}
\end{minipage}
\vfill
\begin{minipage}{0.195\linewidth}
  \centerline{\includegraphics[width=1\linewidth]{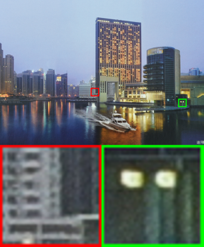}}
  \vspace{-0.03cm}
  \centerline{Ground Truth HR}
  \centerline{(PSNR / SSIM)}
\end{minipage}
\hfill
\begin{minipage}{0.195\linewidth}
  \centerline{\includegraphics[width=1\linewidth]{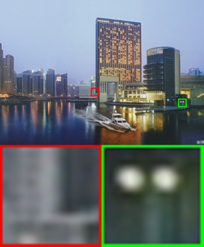}}
  \vspace{-0.03cm}
  \centerline{Interpolated LR}
  \centerline{(28.15 / 0.8113)}
\end{minipage}
\hfill
\begin{minipage}{0.195\linewidth}
  \centerline{\includegraphics[width=1\linewidth]{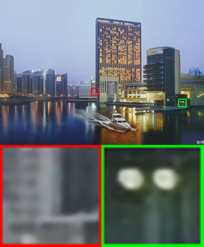}}
  \vspace{-0.03cm}
  \centerline{BicubicSR}
  \centerline{(28.97 / 0.8377)}
\end{minipage}
\hfill
\begin{minipage}{0.195\linewidth}
  \centerline{\includegraphics[width=1\linewidth]{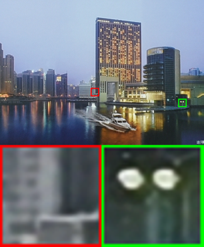}}
  \vspace{-0.03cm}
  \centerline{GaussianSR}
  \centerline{(28.92 / 0.8446)}
\end{minipage}
\hfill
\begin{minipage}{0.195\linewidth}
  \centerline{\includegraphics[width=1\linewidth]{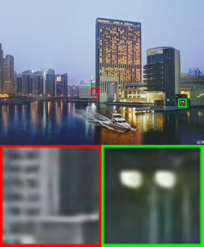}}
  \vspace{-0.03cm}
  \centerline{CameraSR}
  \centerline{(30.26 / 0.8684)}
\end{minipage}
\end{minipage}
\end{center}
\vspace{-0.2cm}
\caption{Visual comparison of SR results under different degradation models in terms of reconstruction accuracy (VDSR \cite{VDSR} network). PSNR and SSIM \cite{SSIM} (the higher, the better) are adopted for evaluation metrics.} 
\label{fig:rset5_acc}
\vspace{-0.35cm}
\end{figure*}

\begin{figure*}[!h]
\fontsize{8}{9.6}\selectfont
\begin{center}
\begin{minipage}{1.0\linewidth}
\begin{minipage}{0.195\linewidth}
  \centerline{\includegraphics[width=1\linewidth]{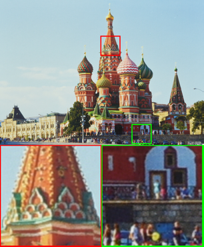}}
  \vspace{-0.03cm}
  \centerline{Ground Truth HR}
  \centerline{(VGG / Ma's metric)}
  \vspace{-0.04cm}
\end{minipage}
\hfill
\begin{minipage}{0.195\linewidth}
  \centerline{\includegraphics[width=1\linewidth]{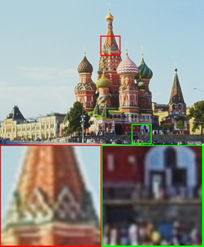}}
  \vspace{-0.03cm}
  \centerline{Interpolated LR}
  \centerline{(0.8543 / 3.58)}
\end{minipage}
\hfill
\begin{minipage}{0.195\linewidth}
  \centerline{\includegraphics[width=1\linewidth]{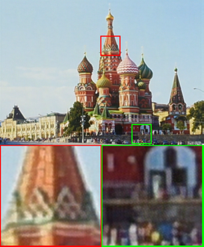}}
  \vspace{-0.03cm}
  \centerline{BicubicSR}
  \centerline{(0.7756 / 5.05)}
\end{minipage}
\hfill
\begin{minipage}{0.195\linewidth}
  \centerline{\includegraphics[width=1\linewidth]{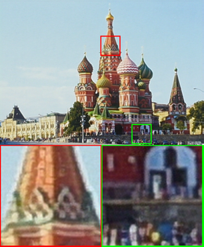}}
  \vspace{-0.03cm}
  \centerline{GaussianSR}
  \centerline{(0.7019 / 6.16)}
\end{minipage}
\hfill
\begin{minipage}{0.195\linewidth}
  \centerline{\includegraphics[width=1\linewidth]{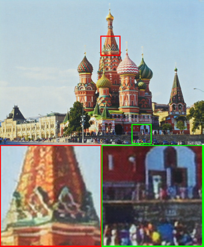}}
  \vspace{-0.03cm}
  \centerline{CameraSR}
  \centerline{(0.4791 / 6.58)}
\end{minipage}
\vfill
\vspace{0.05cm}
\begin{minipage}{0.195\linewidth}
  \centerline{\includegraphics[width=1\linewidth]{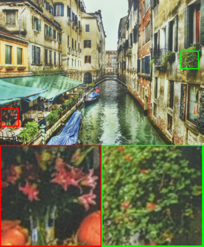}}
  \vspace{-0.03cm}
  \centerline{Ground Truth HR}
  \centerline{(VGG / Ma's metric)}
  \vspace{-0.04cm}
\end{minipage}
\hfill
\begin{minipage}{0.195\linewidth}
  \centerline{\includegraphics[width=1\linewidth]{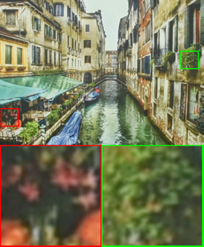}}
  \vspace{-0.03cm}
  \centerline{Interpolated LR}
  \centerline{(0.9654 / 3.58)}
\end{minipage}
\hfill
\begin{minipage}{0.195\linewidth}
  \centerline{\includegraphics[width=1\linewidth]{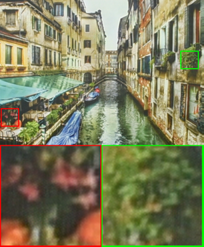}}
  \vspace{-0.03cm}
  \centerline{BicubicSR}
  \centerline{(0.8254 / 4.43)}
\end{minipage}
\hfill
\begin{minipage}{0.195\linewidth}
  \centerline{\includegraphics[width=1\linewidth]{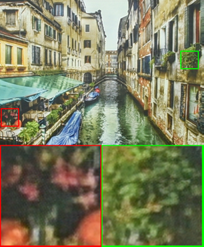}}
  \vspace{-0.03cm}
  \centerline{GaussianSR}
  \centerline{(0.7798 / 5.93)}
\end{minipage}
\hfill
\begin{minipage}{0.195\linewidth}
  \centerline{\includegraphics[width=1\linewidth]{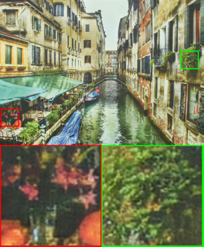}}
  \vspace{-0.03cm}
  \centerline{CameraSR}
  \centerline{(0.6167 / 6.71)}
\end{minipage}
\end{minipage}
\end{center}
\vspace{-0.2cm}
\caption{Visual comparison of SR results under different degradation models in terms of perceptual quality (SRGAN \cite{SRGAN} network). The VGG metric \cite{VGG} (the lower, the better) and the Ma's metric \cite{Ma-Metric} (the higher, the better) are adopted for evaluation.}
\vspace{-0.2cm}
\label{fig:rset5_percep}
\end{figure*}

For perceptual quality, we adopt the SRGAN network \cite{SRGAN} with a combined loss
\begin{equation}
\mathcal{L}_{Comb} = \mathcal{L}_{MSE} + \mathcal{L}_{VGG} + 10e^{-3}\mathcal{L}_{Gen},
\end{equation}
where the VGG loss $\mathcal{L}_{VGG}$ represents the pixel-wise distance in the feature space $\phi(\cdot)$ of a VGG-19 network \cite{VGG}
\begin{equation}
\mathcal{L}_{VGG} = ||\phi(S_\Theta(D(x)))-\phi(x)||_2^2,
\end{equation}
and the generative loss $\mathcal{L}_{Gen}$ is defined based on the probability of a discriminator $\mathcal{D}_{\Theta'}(\cdot)$ as
\begin{equation}
\mathcal{L}_{Gen} = -log\mathcal{D}_{\Theta'}(S_\Theta(D(x))),
\end{equation}
where $\mathcal{D}_{\Theta'}(\cdot)$ denotes the probability that a reconstructed image $S_\Theta(D(X))$ is a natural one. The generative component $S_\Theta(\cdot)$ and the discriminator $\mathcal{D}_{\Theta'}(\cdot)$ are trained in an adversarial manner \cite{GAN}. 

\begin{figure}
\begin{center}
\begin{minipage}{1.0\linewidth}
\begin{minipage}{1.0\linewidth}
\begin{minipage}{0.493\linewidth}
  \centerline{\includegraphics[width=1\linewidth]{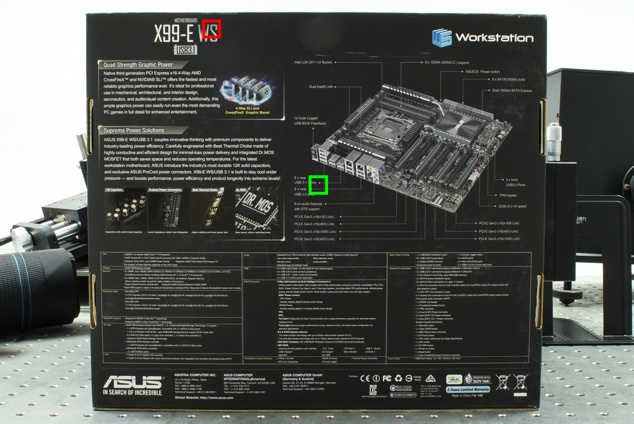}}
  \fontsize{8}{9.6}\selectfont
  \centerline{(a) Captured at 18mm focal length}
\end{minipage}
\hfill
\begin{minipage}{0.493\linewidth}
  \centerline{\includegraphics[width=1\linewidth]{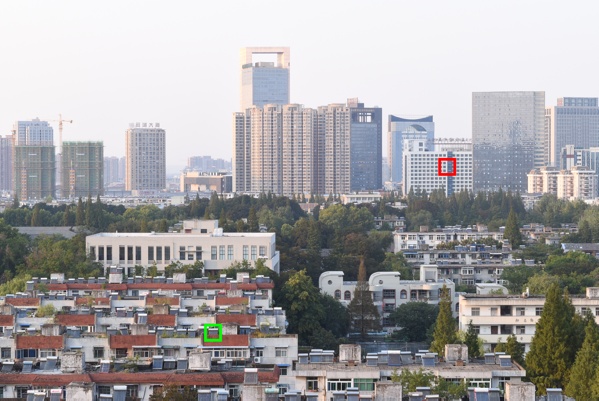}}
  \fontsize{8}{9.6}\selectfont
  \centerline{(b) Captured at 55mm focal length}
\end{minipage}
\end{minipage}
\vfill
\vspace{0.05cm}
\begin{minipage}{1.0\linewidth}
\begin{minipage}{0.24\linewidth}
  \centerline{\includegraphics[width=1\linewidth]{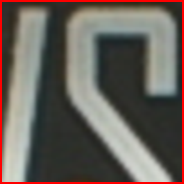}}
\end{minipage}
\hfill
\begin{minipage}{0.24\linewidth}
  \centerline{\includegraphics[width=1\linewidth]{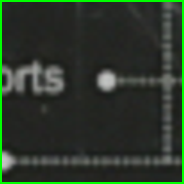}}
\end{minipage}
\hfill
\begin{minipage}{0.24\linewidth}
  \centerline{\includegraphics[width=1\linewidth]{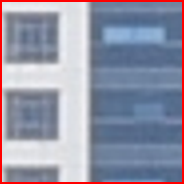}}
\end{minipage}
\hfill
\begin{minipage}{0.24\linewidth}
  \centerline{\includegraphics[width=1\linewidth]{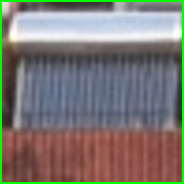}}
\end{minipage}
\vfill
\begin{minipage}{0.24\linewidth}
  \centerline{\includegraphics[width=1\linewidth]{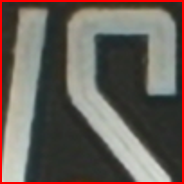}}
\end{minipage}
\hfill
\begin{minipage}{0.24\linewidth}
  \centerline{\includegraphics[width=1\linewidth]{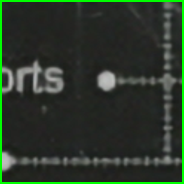}}
\end{minipage}
\hfill
\begin{minipage}{0.24\linewidth}
  \centerline{\includegraphics[width=1\linewidth]{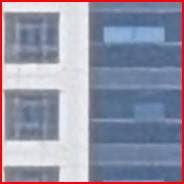}}
\end{minipage}
\hfill
\begin{minipage}{0.24\linewidth}
  \centerline{\includegraphics[width=1\linewidth]{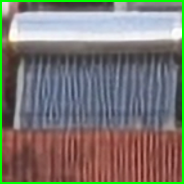}}
\end{minipage}
\vfill
\begin{minipage}{0.24\linewidth}
  \centerline{\includegraphics[width=1\linewidth]{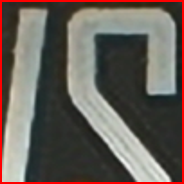}}
\end{minipage}
\hfill
\begin{minipage}{0.24\linewidth}
  \centerline{\includegraphics[width=1\linewidth]{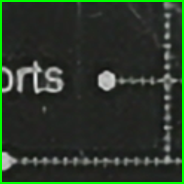}}
\end{minipage}
\hfill
\begin{minipage}{0.24\linewidth}
  \centerline{\includegraphics[width=1\linewidth]{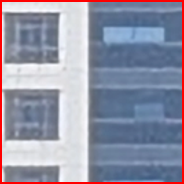}}
\end{minipage}
\hfill
\begin{minipage}{0.24\linewidth}
  \centerline{\includegraphics[width=1\linewidth]{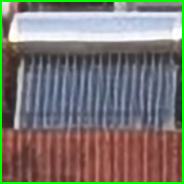}}
\end{minipage}
\vfill
\begin{minipage}{0.24\linewidth}
  \centerline{\includegraphics[width=1\linewidth]{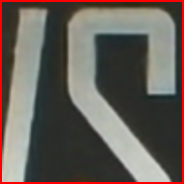}}
\end{minipage}
\hfill
\begin{minipage}{0.24\linewidth}
  \centerline{\includegraphics[width=1\linewidth]{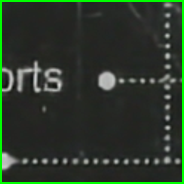}}
\end{minipage}
\hfill
\begin{minipage}{0.24\linewidth}
  \centerline{\includegraphics[width=1\linewidth]{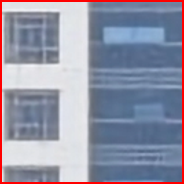}}
\end{minipage}
\hfill
\begin{minipage}{0.24\linewidth}
  \centerline{\includegraphics[width=1\linewidth]{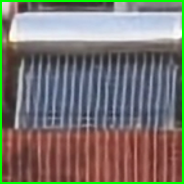}}
\end{minipage}
\vfill
\vspace{0.12cm}
\fontsize{8}{9.6}\selectfont
  \centerline {(c) From top to bottom: Interp. LR, BicubicSR, GaussianSR, CameraSR}
\end{minipage}
\end{minipage}
\end{center}
\vspace{-0.1cm}
\caption{Visual comparison of SR results on images captured by Nikon D5500. SR models are trained on City100 using the VDSR \cite{VDSR} network.}
\vspace{-0.10cm}
\label{fig:real_world_acc1}
\end{figure}

\begin{figure}
\begin{center}
\begin{minipage}{1.0\linewidth}
\fontsize{8}{9.6}\selectfont
\begin{minipage}{1.0\linewidth}
  \centerline{\includegraphics[width=1\linewidth]{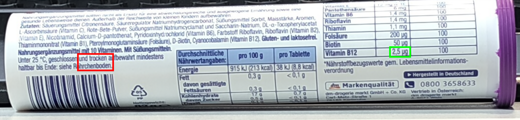}}
\end{minipage}
\vfill
\begin{minipage}{0.325\linewidth}
  \centerline{\includegraphics[width=1\linewidth]{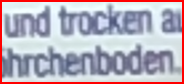}}
\end{minipage}
\hfill
\begin{minipage}{0.325\linewidth}
  \centerline{\includegraphics[width=1\linewidth]{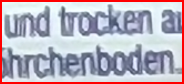}}
\end{minipage}
\hfill
\begin{minipage}{0.325\linewidth}
  \centerline{\includegraphics[width=1\linewidth]{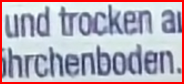}}
\end{minipage}
\vfill
\begin{minipage}{0.325\linewidth}
  \centerline{\includegraphics[width=1\linewidth]{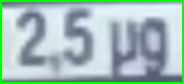}}
\end{minipage}
\hfill
\begin{minipage}{0.325\linewidth}
  \centerline{\includegraphics[width=1\linewidth]{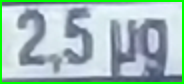}}
\end{minipage}
\hfill
\begin{minipage}{0.325\linewidth}
  \centerline{\includegraphics[width=1\linewidth]{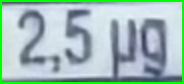}}
\end{minipage}
\vfill
\vspace{0.05cm}
\begin{minipage}{0.325\linewidth}
  \centerline{Digital Zoom}
\end{minipage}
\hfill
\begin{minipage}{0.325\linewidth}
  \vspace{-0.08cm}
  \centerline{BicubicSR}
\end{minipage}
\hfill
\begin{minipage}{0.325\linewidth}
  \vspace{-0.08cm}
  \centerline{CameraSR}
\end{minipage}
\vspace{0.05cm}
\begin{minipage}{1.0\linewidth}
  \centerline{\includegraphics[width=1\linewidth]{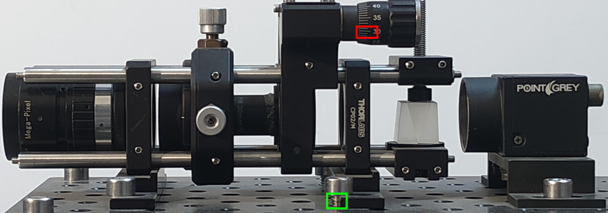}}
\end{minipage}
\vfill
\vspace{-0.05cm}
\begin{minipage}{0.325\linewidth}
  \centerline{\includegraphics[width=1\linewidth]{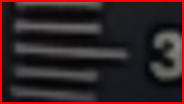}}
\end{minipage}
\hfill
\begin{minipage}{0.325\linewidth}
  \centerline{\includegraphics[width=1\linewidth]{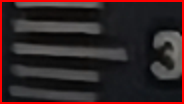}}
\end{minipage}
\hfill
\begin{minipage}{0.325\linewidth}
  \centerline{\includegraphics[width=1\linewidth]{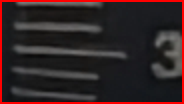}}
\end{minipage}
\vfill
\begin{minipage}{0.325\linewidth}
  \centerline{\includegraphics[width=1\linewidth]{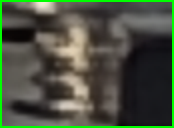}}
\end{minipage}
\hfill
\begin{minipage}{0.325\linewidth}
  \centerline{\includegraphics[width=1\linewidth]{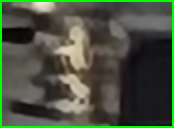}}
\end{minipage}
\hfill
\begin{minipage}{0.325\linewidth}
  \centerline{\includegraphics[width=1\linewidth]{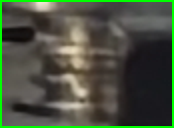}}
\end{minipage}
\vfill
\vspace{0.1cm}
\begin{minipage}{0.325\linewidth}
  \centerline{Digital Zoom}
\end{minipage}
\hfill
\begin{minipage}{0.325\linewidth}
  \vspace{-0.08cm}
  \centerline{BicubicSR}
\end{minipage}
\hfill
\begin{minipage}{0.325\linewidth}
  \vspace{-0.08cm}
  \centerline{CameraSR}
\end{minipage}
\end{minipage}
\end{center}
\vspace{-0.1cm}
\caption{Visual comparison of SR results on images captured by iPhone X. SR models are trained on the smartphone version of City100 using the VDSR \cite{VDSR} network.}
\vspace{-0.15cm}
\label{fig:real_world_acc2}
\end{figure}

Then, we train two sets of SR networks for $D_{Bic}(\cdot)$, $D_{Gau}(\cdot)$, and $D_{RV}(\cdot)$ based on City100, respectively. All other hyper-parameters except the degradation model are kept the same to eliminate the influence of them. The quantitative results evaluated on PSNR are shown in Fig.~\ref{fig:comparison_lr_sr}(b), where both BicubicSR and GaussianSR have a notable performance gap (i.e., about 1.3 dB in average on the test set) compared with CameraSR.
For GaussianSR, we evaluate two settings at the red points in Fig.~\ref{fig:comparison_lr_sr}(a) and report the better one.
Detailed quantitative results are listed in Table~\ref{tab}. The corresponding visual comparisons are conducted in Figs.~\ref{fig:rset5_acc} and \ref{fig:rset5_percep} for VDSR \cite{VDSR} and SRGAN \cite{SRGAN} respectively, which again validates the significantly improved SR results achieved by CameraSR. More results for comparison can be found in the supplementary document.

\vspace{-0.05cm}
\section{Experiments}
\vspace{-0.05cm}
\label{sec:camsr}

While the above analysis clearly demonstrates the importance of degradation modeling for the resolution enhancement of realistic imaging systems, it is not so surprising that CameraSR outperforms BicubicSR and GaussianSR since it directly learns the R-V degradation from City100.
In this section, we show extensive SR results to demonstrate the generalizability of CameraSR (still trained on City100) to real-world scenes that are drastically different from City100 in content and even captured with different devices. Still, BicubicSR and GaussianSR are adopted for comparisons, in terms of reconstruction accuracy and perceptual quality.

\begin{figure*}[!h]
\begin{center}
\begin{minipage}{1.0\linewidth}
\begin{minipage}{0.495\linewidth}
\begin{minipage}{0.336\linewidth}
  \vspace{-0.025cm}
  \centerline{\includegraphics[width=1\linewidth]{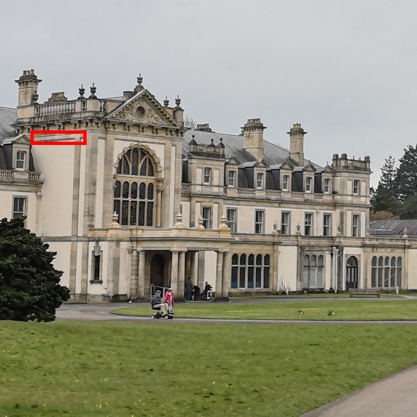}}
\end{minipage}
\hfill
\hspace{-0.33cm}
\begin{minipage}{0.688\linewidth}
  \centerline{\includegraphics[width=0.9\linewidth]{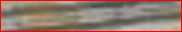}}
\vfill
\vspace{0.02cm}
  \centerline{\includegraphics[width=0.9\linewidth]{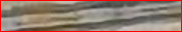}}
\vfill
\vspace{0.02cm}
  \centerline{\includegraphics[width=0.9\linewidth]{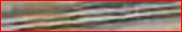}}
\end{minipage}
\vfill
\vspace{0.1cm}
\fontsize{8}{9.6}\selectfont
\ \ (a) Huawei P20. Top to bottom: Interp. LR, BicubicSR, and  CameraSR.
\end{minipage}
\hfill
\hspace{-0.3cm}
\begin{minipage}{0.512\linewidth}
\begin{minipage}{0.226\linewidth}
  \centerline{\includegraphics[width=1\linewidth]{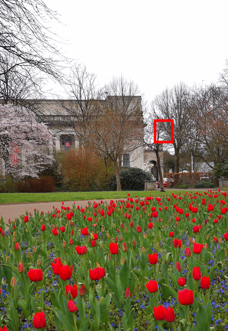}}
\end{minipage}
\hfill
\hspace{-0.15cm}
\begin{minipage}{0.25\linewidth}
  \centerline{\includegraphics[width=1\linewidth]{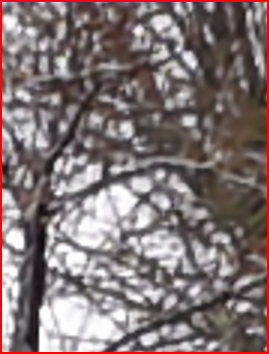}}
\end{minipage}
\hfill
\hspace{-0.15cm}
\begin{minipage}{0.25\linewidth}
  \centerline{\includegraphics[width=1\linewidth]{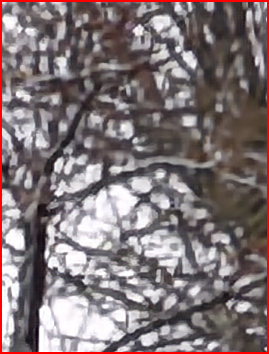}}
\end{minipage}
\hfill
\hspace{-0.15cm}
\begin{minipage}{0.25\linewidth}
  \centerline{\includegraphics[width=1\linewidth]{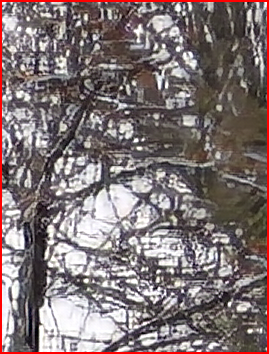}}
\end{minipage}
\vfill
\vspace{0.1cm}
\begin{minipage}{0.226\linewidth}
  \fontsize{8}{9.6}\selectfont
  \centerline{(b) Huawei P20}
\end{minipage}
\hfill
\hspace{-0.15cm}
\begin{minipage}{0.25\linewidth}
  \fontsize{8}{9.6}\selectfont
  \centerline{Interpolated LR}
\end{minipage}
\hfill
\hspace{-0.15cm}
\begin{minipage}{0.25\linewidth}
  \fontsize{8}{9.6}\selectfont
  \vspace{-0.04cm}
  \centerline{BicubicSR}
\end{minipage}
\hfill
\hspace{-0.15cm}
\begin{minipage}{0.25\linewidth}
  \fontsize{8}{9.6}\selectfont
  \vspace{-0.04cm}
  \centerline{CameraSR}
\end{minipage}
\end{minipage}
\vfill
\vspace{0.1cm}

\begin{minipage}{0.1435\linewidth}
  \centerline{\includegraphics[width=1\linewidth]{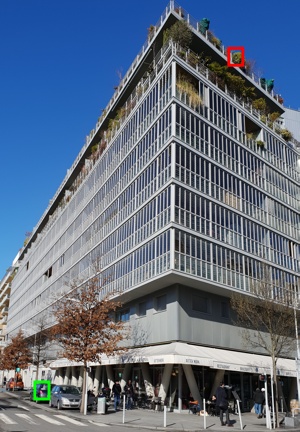}}
\end{minipage}
\hfill
\hspace{-0.1cm}
\begin{minipage}{0.141\linewidth}
  \centerline{\includegraphics[width=1\linewidth]{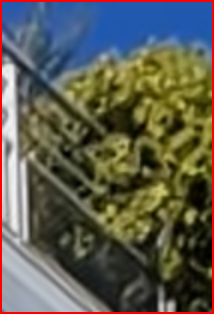}}
\end{minipage}
\hfill
\hspace{-0.1cm}
\begin{minipage}{0.141\linewidth}
  \centerline{\includegraphics[width=1\linewidth]{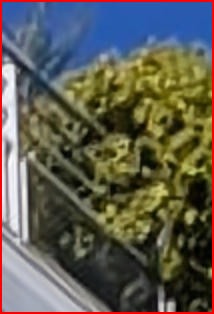}}
\end{minipage}
\hfill
\hspace{-0.1cm}
\begin{minipage}{0.141\linewidth}
  \centerline{\includegraphics[width=1\linewidth]{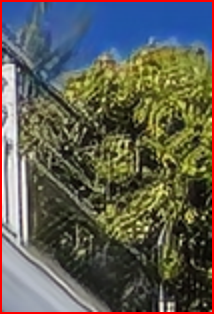}}
\end{minipage}
\hfill
\hspace{-0.1cm}
\begin{minipage}{0.1343\linewidth}
  \centerline{\includegraphics[width=1\linewidth]{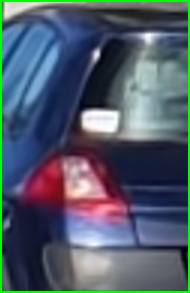}}
\end{minipage}
\hfill
\hspace{-0.1cm}
\begin{minipage}{0.1343\linewidth}
  \centerline{\includegraphics[width=1\linewidth]{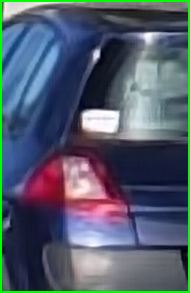}}
\end{minipage}
\hfill
\hspace{-0.1cm}
\begin{minipage}{0.1343\linewidth}
  \centerline{\includegraphics[width=1\linewidth]{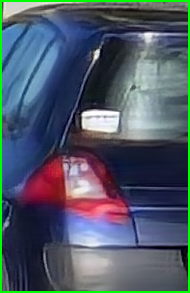}}
\end{minipage}
\vfill
\vspace{0.05cm}
\begin{minipage}{0.1435\linewidth}
  \vspace{0.06cm}
  \fontsize{8}{9.6}\selectfont
  \centerline{(c) Samsung S9}
\end{minipage}
\hfill
\hspace{-0.1cm}
\begin{minipage}{0.141\linewidth}
  \vspace{0.075cm}
  \fontsize{8}{9.6}\selectfont
  \centerline{Interpolated LR}
\end{minipage}
\hfill
\hspace{-0.1cm}
\begin{minipage}{0.141\linewidth}
  \fontsize{8}{9.6}\selectfont
  \centerline{BicubicSR}
\end{minipage}
\hfill
\hspace{-0.1cm}
\begin{minipage}{0.141\linewidth}
\fontsize{8}{9.6}\selectfont
  \centerline{CameraSR}
\end{minipage}
\hfill
\hspace{-0.1cm}
\begin{minipage}{0.1343\linewidth}
  \fontsize{8}{9.6}\selectfont
  \vspace{0.075cm}
  \centerline{Interpolated LR}
\end{minipage}
\hfill
\hspace{-0.1cm}
\begin{minipage}{0.1343\linewidth}
  \fontsize{8}{9.6}\selectfont
  \centerline{BicubicSR}
\end{minipage}
\hfill
\hspace{-0.1cm}
\begin{minipage}{0.1343\linewidth}
  \fontsize{8}{9.6}\selectfont
  \centerline{CameraSR}
\end{minipage}
\end{minipage}
\end{center}
\vspace{-0.1cm}
\caption{Visual comparison of SR results on images captured by Huawei P20 and Samsung S9 smartphone cameras. SR models are trained on the iPhone X version of City100 using the VDSR \cite{VDSR} network for (a) and SRGAN \cite{SRGAN} network for (b) and (c), respectively.}
\label{fig:ipx_hw2}
\vspace{-0.1cm}
\end{figure*}

\subsection{Advanced digital zoom}
Recall that our main goal is to alleviate the R-V tradeoff or even break the physical zoom ratio of an optical lens in realistic imaging systems, we now demonstrate that CameraSR achieves this goal. As shown in Fig.~\ref{fig:real_world_acc1}(a), given an image captured by a DSLR camera at the focal length of 18mm, CameraSR effectively super-resolves its details, which can be viewed as alleviating the R-V tradeoff of the camera lens (i.e., resolution and FoV are now obtained at the same time).
Meanwhile, when the zoom lens of the same DSLR camera reaches its maximum magnification at the focal length of 55mm, CameraSR is capable of further enhancing the resolution of the captured image, as shown in Fig.~\ref{fig:real_world_acc1}(b). Similarly in Fig.~\ref{fig:real_world_acc2}, for a smartphone camera with a fixed focal lens, CameraSR serves as an advanced digital zoom tool, which significantly enhances the imaging quality compared with the built-in digital zoom function. The examples in Fig.~\ref{fig:real_world_acc1}(b) and Fig.~\ref{fig:real_world_acc2} can be viewed as breaking the physical limit of zoom ratio.

\subsection{Generalizability}
\label{sec:generalization}

Besides the significant improvement of SR performance, our proposed CameraSR also has a favorable generalization capability in terms of both content and device. For the content generalization, recall that the City100 dataset is captured under an indoor environment with a single category of subjects (i.e., postcard), yet the CameraSR model trained on City100 performs well in both indoor and outdoor environments with diverse subjects, as demonstrated in Figs.~\ref{fig:real_world_acc1}, \ref{fig:real_world_acc2}, \ref{fig:ipx_hw2}.
For the device generalization, as shown in Fig.~\ref{fig:ipx_hw2}, the CameraSR model trained on the iPhone X version of City100 can be readily applied to different smartphones such as Huawei P20 and Samsung S9. More results for the generalization from Nikon to Canon DSLR cameras are shown in the supplementary document.

\section{Conclusion and Discussion}
\label{sec:conclusion}

In this paper, we investigate SR from the perspective of camera lenses, named as CameraSR, which models the R-V degradation in realistic imaging systems. With the proposed data acquisition strategies, we build a City100 dataset to characterize the R-V degradation in representative DSLR and smartphone cameras. Based on City100, we analyze the disadvantage of the commonly used synthetic degradation models and validate CameraSR as a practical solution to boost the performance of existing SR methods. Due to its favorable generalization capability, CameraSR could find a wide application as an advanced digital zoom tool in realistic imaging systems. Especially, besides the enhancement of natural images, we believe CameraSR has a great value for biomedical imaging with microscopes, where the resolution enhancement is essential for scientific observation.

Despite the promising preliminary results, there are still some real-world conditions that have not been considered in this paper. In terms of the LR observation, we consider a relatively ideal condition without noise. Yet the influence of noise is inevitable, especially in the smartphone imaging systems with small sensors. It is thus worth to jointly investigate the R-V degradation and noise to further promote the robustness of CameraSR.
Besides single image SR discussed in this paper, the R-V degradation can be generalized to burst image SR, where a sequence of LR images are captured using the burst shooting mode to exploit the underlying information from the sub-pixel motion for a better HR reconstruction.
Moreover, beyond the prior learned from external examples, the proposed CameraSR can be further extended for self-similarity based methods to utilize the inherent recurrence, by numerically estimating the R-V degradation kernel based on City100. The above extensions are considered as our future work.

\vspace{-0.05cm}
\section*{Acknowledgement}
\vspace{-0.08cm}
We acknowledge funding from National Key R\&D Program of China under Grant 2017YFA0700800, and Natural Science Foundation of China (NSFC) under Grants 61671419, 61425026, 61622211 and 61620106009.

{\small
\bibliographystyle{ieee}
\bibliography{camsr}
}

\end{document}